\documentclass[11pt,a4paper]{article}

\usepackage[table]{xcolor} %color in tables 
\usepackage{pgfplots}
\pgfplotsset{compat=1.7}

\usepackage[hyperref]{acl2017}
\usepackage{times}
\usepackage{latexsym}

\usepackage{url}

\usepackage[framemethod=tikz]{mdframed}
\usepackage{framed}     %frame over a text
\usepackage{subcaption}
\usepackage[utf8]{inputenc} %UTF8
\usepackage[T1]{fontenc} %UTF8
\usepackage{tabularx,lipsum} %fit the table to column width
\usepackage{enumitem}
\usepackage{times}
\usepackage{url}
\usepackage{color,soul}
\usepackage[colorinlistoftodos]{todonotes}
\usepackage{csvsimple} %read csv file and import in a table.
\usepackage{amssymb} % adding checkmark
\usepackage{multirow} %multi rows in tables
\usepackage{tikz}
\usetikzlibrary{positioning}

\definecolor{LightGray}{gray}{0.85}
\definecolor{Gray}{gray}{0.75}

\newcounter{example}

\def \corpus {ConcoDisco}

\title{Improving Discourse Relation Projection to \\ Build Discourse Annotated Corpora}

\aclfinalcopy % Uncomment this line for the final submission
\author{Majid Laali \hspace{2cm} Leila Kosseim\\
  Department of Computer Science and Software Engineering \\
  Concordia University, Montreal, Quebec, Canada \\
  {\tt \{m\_laali, kosseim\}@encs.concordia.ca} \\
  }

\date{}

\begin{document}

\maketitle

\begin{abstract}

    The naive approach to annotation projection is not effective to project discourse annotations from one language to another because implicit discourse relations are often changed to explicit ones and vice-versa in the translation. In this paper, we propose a novel approach based on the intersection between statistical word-alignment models to identify unsupported discourse annotations. This approach identified 65\% of the unsupported annotations in the English-French parallel sentences from Europarl. By filtering out these unsupported annotations, we induced the first PDTB-style discourse annotated corpus for French from Europarl. We then used this corpus to train a classifier to identify the discourse-usage of French discourse connectives and show a 15\% improvement of F1-score compared to the classifier trained on the non-filtered annotations.
   
\end{abstract}

%%%%%%%%%%%%%%%%%%%%%%%%%%%%%%%%%%%%%%%%%%%%%%%%%%%%%%%%%%%%%%%%%%%%%%%%%%%%%%%%%%%%%%%%%%
\section{Introduction}
%%%%%%%%%%%%%%%%%%%%%%%%%%%%%%%%%%%%%%%%%%%%%%%%%%%%%%%%%%%%%%%%%%%%%%%%%%%%%%%%%%%%%%%%%%

The Penn Discourse Treebank (PDTB) \cite{prasad08} is one of the most successful projects aimed at the development of discourse annotated corpora. Following the predicate-argument approach of the D-LTAG framework \cite{webber03}, the PDTB associates discourse relations (DRs) to lexical elements, so-called \textit{discourse connectives (DCs)}. More specifically, DRs between two text spans (so-called \textit{discourse arguments}) are triggered by either lexical elements (or \textit{explicit DCs}) such as \textit{however, because} or without any lexical element and are inferred by the reader. If a DR is inferred by the reader, annotators of the PDTB inserted an inferred DC which conveys the same DR between the text spans  (or \textit{implicit DCs}). As a result of this annotation schema, DCs were heavily used to annotate DRs in the PDTB. 

Manually constructing PDTB-style discourse annotated corpora is expensive, both in terms of time and expertise. As a result, such corpora are only available for a limited number of languages. 

Annotation projection is an effective approach to quickly build initial discourse treebanks using parallel sentences. The main assumption of annotation projection is that because parallel sentences are a translation of each other, semantic annotations can be projected from one side onto the other side of parallel sentences. However, this assumption does not always hold for the projection of discourse annotations because the realization of DRs can change during the translation. More specifically, although parallel sentences may convey the same DR, implicit DRs are often changed to explicit DRs and vice versa \cite{zufferey12,meyer13,cartoni13,zufferey15,zufferey16}. In this paper, we focus on the case when an explicit DR is changed to an implicit one, hence explicit DCs are removed during the translation process. Example~(\ref{ex:mais}) shows parallel sentences where the French DC \textit{mais}\footnote{Free translation: \textit{but}} has been dropped in the English translation. 

{\small
\begin{enumerate}[label=(\arabic*)]
    \setcounter{enumi}{\theexample}
    \refstepcounter{example}
    \label{ex:mais}
    \item FR:  \textit{Comme tout le monde dans cette Assemblée, j'aspire à cet espace de liberté, de justice et de sécurité, \textbf{mais} je ne veux pas qu'il débouche sur une centralisation à outrance, le chaos et la confusion.}\\
    EN: \textit{Like everybody in this House, I want freedom, justice and security. I do not want to see these degenerate into over-centralisation, chaos and confusion.}
\end{enumerate}
}

According to \newcite{meyer13}, up to 18\% of explicit DRs are changed to implicit ones in the English/French portion of the newstest2010+2012 dataset \cite{callison-burch10,callison-burch12}. Because no counterpart translation exists for the new explicit DCs, it is difficult to reliably annotate them and any induced annotation would be unsupported. 

To address this problem, we propose a novel method based on the intersection between statistical word-alignment models to identify unsupported annotations. We experimented with English-French parallel texts from Europarl \cite{koehn05} and projected discourse annotations from English texts onto French texts. Our approach identified 65\% of unsupported discourse annotations.
%, which is significantly better than naive annotation projection. 
Using our approach, we then induced the first PDTB-style discourse annotated corpus for French\footnote{The corpus is available at \url{https://github.com/mjlaali/Europarl-ConcoDisco}} and used it to train a classifier that identifies the discourse usage of French DCs. Our results show that  filtering unsupported annotations improves the relative F1-score of the classifier by 15\%.

%%%%%%%%%%%%%%%%%%%%%%%%%%%%%%%%%%%%%%%%%%%%%%%%%%%%%%%%%%%%%%%%%%%%%%%%%%%%%%%%%%%%%%%%%%
\section{Related Work}
\label{sec:related-work}
%%%%%%%%%%%%%%%%%%%%%%%%%%%%%%%%%%%%%%%%%%%%%%%%%%%%%%%%%%%%%%%%%%%%%%%%%%%%%%%%%%%%%%%%%%

Annotation projection has been widely used in the past to build natural language applications and resources. It has been applied for POS tagging \cite{yarowsky01}, word sense disambiguation \cite{bentivogli05} and dependency parsing \cite{tiedemann15} and more recently, for inducing discourse resources \cite{versley10,laali14,hidey16}. These works implicitly assume that linguistic annotations can be projected from one side onto the other side in parallel sentences; however, this may not always be the case. In this work, we pay special attention to parallel sentences for which this assumption does not hold and therefore, the projected annotations are not supported.  

In the context of DR projection, the realization of DRs may be changed from explicit to implicit during the translation, hence explicit DCs are dropped in the translation process \cite{zufferey12,meyer13,cartoni13,zufferey15,zufferey16}. To extract dropped DCs, authors either manually annotate parallel sentences \cite{zufferey12,zufferey15,zufferey16} or use a heuristic based approach using a dictionary \cite{meyer13,cartoni13} to verify the translation of DCs proposed by statistical word alignment models such as IBM models \cite{brown93}.
%This phenomenon has been studied in the context of machine translation \cite{meyer13} and DC meaning \cite{zufferey12}. 
%For example, \newcite{meyer13} showed that 18\% of explicit relations are changed to implicit ones in the English/French portion of the newstest2010+2012 dataset \cite{callison-burch10,callison-burch12}. 
%On the other hand, \newcite{zufferey12} manually annotated 1,076 causal DCs to identify their translations and found out 49 of them are dropped during the translation. 
%In this work, we also propose an approach to identify dropped DCs during the translation. 
%in order to identify parallel sentences where the explicit DRs have been changed to implicit ones. 
In contrast to previous works, our approach automatically identifies dropped DCs by intersecting statistical word-alignments without using any additional resources such as a dictionary. 

Note that, because DRs are semantic and rhetorical in nature, even though explicit DCs may be removed during the translation process, we assume that DRs are preserved during the translation process. Therefore, the DRs should, in principle, be transferred from the source language to the target language. Although this assumption is not directly addressed in previous work, it has been implicitly used by many (e.g. \cite{hidey16,laali14,cartoni13,popescu-belis12,meyer11,versley10,prasad10}). 
 %However, we do not assume the realization of DR are the same in both sentences of a parallel sentence.

As a by-product of this work, 
we also generated a PDTB-style discourse annotated corpus for French. Currently, there exist two publicly available  discourse annotated corpora for French: \textit{The French Discourse Treebank (FDTB)} \citep{danlos15} and \textit{ANNODIS} \citep{afantenos12}. The FDTB corpus contains more than 10,000 instances of French discourse connectives annotated as \textit{discourse-usage}. However, to date, French discourse connectives have not been annotated with DRs. On the other hand, while \textit{ANNODIS} contains DRs, the relations are not associated to DCs. Moreover, the size of the corpus is small and only contains 3355 relations.

%LEXCONN \cite{danlos15} is a manually built lexicon of French DCs containing 343 DCs mapped to an average of 1.3 DRs taken from various sources, but not from the PDTB. The French Discourse Treebank (FDTB) \cite{danlos15} contains more than 10,000 instances of LEXCONN's French DCs annotated as \textit{discourse-usage} in two syntactically annotated corpora: the Sequoia Treebank \cite{candito12} and the French Treebank (FTB) \cite{abeille00}. However, to date, there exists no corpus where French DCs are annotated with PDTB DRs.

%%%%%%%%%%%%%%%%%%%%%%%%%%%%%%%%%%%%%%%%%%%%%%%%%%%%%%%%%%%%%%%%%%%%%%%%%%%%%%%%%%%%%%%%%%
\section{Methodology}

%%%%%%%%%%%%%%%%%%%%%%%%%%%%%%%%%%%%%%%%%%%%%%%%%%%%%%%%%%%%%%%%%%%%%%%%%%%%%%%%%%%%%%%%%%
\subsection{Corpus Preparation}
\label{sec:corpus-preparation}
%%%%%%%%%%%%%%%%%%%%%%%%%%%%%%%%%%%%%%%%%%%%%%%%%%%%%%%%%%%%%%%%%%%%%%%%%%%%%%%%%%%%%%%%%%

For our experiment, we have used the English-French part of the Europarl corpus \cite{koehn05} which contains around two million parallel sentences and around 50 millions words in each side. To prepare this dataset for our experiment, we used the CLaC discourse parser~\cite{laali16} to identify English DCs and the DR that they signal. The CLaC parser has been trained on Section 02-20 of the PDTB and can disambiguate the usage of the 100 English DCs listed in the PDTB with an F1-score of 0.90 and label them with their PDTB relation with an F1-score of 0.76 when tested on the blind test set of the CoNLL 2016 shared task~\cite{xue16}. 
This parser was used because its performance is very close to that of the state of the art~\cite{oepen16} (i.e. 0.91 and 0.77 respectively), but is more efficient at running time than~\newcite{oepen16}. 
Note that since the CoNLL~2016 blind test set was extracted from Wikipedia and its domain and genre differ significantly from the PDTB, the 0.90 and 0.76 F1-scores of the CLaC parser can be considered as an estimation of its performance on texts with a different domain such as Europarl. 

%%%%%%%%%%%%%%%%%%%%%%%%%%%%%%%%%%%%%%%%%%%%%%%%%%%%%%%%%%%%%%%%%%%%%%%%%%%%%%%%%%%%%%%%%%
\subsection{Discourse Annotation Projection}
%%%%%%%%%%%%%%%%%%%%%%%%%%%%%%%%%%%%%%%%%%%%%%%%%%%%%%%%%%%%%%%%%%%%%%%%%%%%%%%%%%%%%%%%%%

\newcommand{\specialcell}[2][c]{%
  \begin{tabular}[#1]{@{}l@{}}#2\end{tabular}}

\newcommand{\explbl}[1]{ex:#1}

\begin{table*}[ht]
    \resizebox{\textwidth}{!}{   
    %\rowcolors{2}{LightGray}{white}
    
    \begin{tabularx}{1.2\textwidth}{|r|X|X|p{4cm}|}
        \hline
        \rowcolor{Gray}
        \textbf{\#} & \textbf{French} & \textbf{English} & \textbf{Projected Annotation} 
        \begin{filecontents*}{ap-examples.csv}
id;EnLeft;aligned;EnRight;EnSense;FrLeft;FrDc;FrRight;Tag;Rel;Inc
1;The Member States must ;also/\allowbreak DU/\allowbreak \textsc{conjunction};bear in mind their responsibility.;conjunction;Les États membres ont\ ;aussi\ ;leur part de responsabilité dans ce domaine et ils ne doivent pas l'oublier.;DU;\textsc{/conjunction}; included in corpus
2;When I speak of optimum utilisation, I am referring ;both/NDU;to the national and regional levels.;none;Et quand je parle d'utilisation optimale, j'évoque\  ;aussi\ ;bien le niveau national que le niveau régional.;NDU; ; included in corpus
3;The conclusion is that we must make the case for guidelines to be broad, indicative and flexible to assist our programme managers and fund-users and to get the maximum potential out of our new fields of regeneration.;;;reason;Pour conclure, je dirai que nous devons faire en sorte que les lignes directrices soient larges, indicatives et souples,\ ; afin d';aider nos gestionnaires de programmes et les utilisateurs des crédits et de valoriser au mieux les potentialités de nos nouveaux domaines de régénération.;None; ; not included in corpus
4;You will tell me that situations of growth or shortage do not affect everyone alike.;;;none;Vous me direz que la croissance ou la pénurie, ce n'est pas ;\ pour\ ; tout le monde.;None; ; not included in corpus
\end{filecontents*}
\csvreader[head to column names,/csv/separator=semicolon]{ap-examples.csv}{}
        {\\\hline  \refstepcounter{example} (\theexample) \label{\explbl{\theexample}} & \textit{\FrLeft\textbf{\FrDc}\FrRight} & \textit{\EnLeft\ \textbf{\aligned}\ \EnRight} & \specialcell[t]{ \Tag \Rel \\ $\Rightarrow$ \Inc }}
        \\ \hline
    \end{tabularx}
    }
    \caption{Examples of discourse connective annotation projection in parallel sentences. French candidate DCs and their correct English translation are in bold face\textsuperscript{\ref{fnote:examples}}. }
    \label{tbl:annotation-projection}
\end{table*}

Once the English side of Europarl was parsed with the CLaC parser, to project these discourse annotations from the English texts onto French texts, we first identified all occurrences of the 371 French DCs listed in LEXCONN \cite{roze12}, in the French side of the parallel texts and marked them as French candidate DCs. Then, we looked in the English translation of the French candidate DCs (see Section~\ref{sec:word-alignment-models}) and we divided the candidates into two categories with respect to their translation: (1)~\textit{supported candidates} (see Section~\ref{sec:supported}), and (2)~\textit{unsupported candidates} (see Section~\ref{sec:unsupported}).

%%%%%%%%%%%%%%%%%%%%%%%%%%%%%%%%%%%%%%%%%%%%%%%%%%%%%%%%%%%%%%%%%%%%%%%%%%%%%%%%%%%%%%%%%%
\subsubsection{Identifying the Translations of Candidate DCs}
\label{sec:word-alignment-models}
%%%%%%%%%%%%%%%%%%%%%%%%%%%%%%%%%%%%%%%%%%%%%%%%%%%%%%%%%%%%%%%%%%%%%%%%%%%%%%%%%%%%%%%%%%

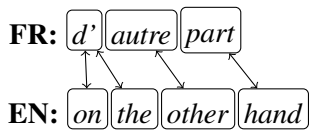
\begin{figure}[htbp]
    \centering
    \begin{tikzpicture}[
            rounded corners=2pt,
            inner sep=2pt,
            text height=1em, 
            node distance=1pt,
            %scale=0.6, every node/.style={scale=0.6}
            ]
        \node [](en) {\textbf{FR:}};
        \node [draw,right=of en](en0) {\textit{d'}};
        \node [draw,right=of en0](en1) {\textit{autre}};
        \node [draw,right=of en1](en2) {\textit{part}};
        
        \node [below=0.5cm of en](fr) {\textbf{EN:}};
        \node [draw,right=of fr](fr0) {\textit{on}};
        \node [draw,right=of fr0](fr1) {\textit{the}};
        \node [draw,right=of fr1](fr2) {\textit{other}};
        \node [draw,right=of fr2](fr3) {\textit{hand}};

        \draw [<->] (en0) -- (fr0);
        \draw [<->] (en0) -- (fr1);
        \draw [<->] (en1) -- (fr2);
        \draw [<->] (en2) -- (fr3);
    \end{tikzpicture}
    \caption{Word-alignment for the French DC \textit{d'autre part}.}
    \label{fig:word-alignment}
\end{figure}

To automatically identify the translation of French candidate DCs, we used statistical world-alignment models. More specifically, we concatenated all the English words that were aligned with each word of the French candidate DCs and considered this concatenation as their English translation. For example, Figure~\ref{fig:word-alignment} shows word-alignments for the French DC \textit{d'autre part} where the alignment model found a 1:2 alignment between \textit{d'} and \textit{on the} then three 1:1 alignments. In this case, the English translation of \textit{d'autre part} will be considered to be \textit{on the other hand}. 

To align English and French words, we used the Moses statistical machine translation system \cite{koehn07}. As part of its translation model, Moses can use a variety of statistical word-alignment models. While previous works only experimented with the \textit{Grow-diag} model \cite{versley10,tiedemann15}, in this work we experimented with different models to identify their effect on the annotation projection task. For our experiment, we trained an IBM 4 word-alignment model in both directions and generated two word-alignments:

\begin{enumerate}
    \item \textit{Direct} word-alignment which includes word-alignments when the source language is set to French and the target language is set to English.
    \item \textit{Inverse} word-alignment which is learned in the reverse direction of \textit{Direct} word-alignment (i.e. the source language is English and the target language is French). 
\end{enumerate}

In addition to these two word-alignments, we also experimented with: 
\begin{enumerate}[resume]
    \item \textit{Intersection} word-alignment which contains alignments that appear in both the \textit{Direct} word-alignment and in the \textit{Inverse} word-alignment. This creates less, but more accurate alignments.
    \item \textit{Grow-diag} word-alignment which expands the \textit{Intersection} word-alignment with the alignments that lie in the union of the \textit{Direct} word-alignment and the \textit{Inverse} word-alignment and that satisfy the heuristic proposed by \newcite{och03}. This heuristic creates more, but less supported alignments.
\end{enumerate}

%%%%%%%%%%%%%%%%%%%%%%%%%%%%%%%%%%%%%%%%%%%%%%%%%%%%%%%%%%%%%%%%%%%%%%%%%%%%%%%%%%%%%%%%%%
\subsubsection{Supported French Candidate DCs}
\label{sec:supported}
%%%%%%%%%%%%%%%%%%%%%%%%%%%%%%%%%%%%%%%%%%%%%%%%%%%%%%%%%%%%%%%%%%%%%%%%%%%%%%%%%%%%%%%%%%
If a French candidate DC has been translated into English in the parallel sentence and has been aligned to English texts, we consider it as a supported candidate and label it according to the annotation of its English translation identified by the word alignments as follows:
 
\begin{enumerate}
    \item \textit{Discourse-Usage (or DU)}: If the English translation was part of a PDTB English DC and was marked by the CLaC discourse parser, then we project the English annotations and assume that the French candidate DC signals the same relation as the English DC. 
    %Therefore, we label the French candidate DC as \textit{DU} and assign it the same DR signalled by the English DC. 
    \item \textit{Non-Discourse-Usage (or NDU)}: If the English translation was not part of a PDTB English DC or was not marked by the CLaC parser, then we project the English NDU label and assume that the French candidate DC is not used in a discourse usage and label it as \textit{NDU}.
\end{enumerate}

%fixed
For example, consider Sentences~(\ref{ex:2}) and (\ref{ex:3}) in Table~\ref{tbl:annotation-projection}. In Sentence~(\ref{ex:2}), \textit{aussi} is translated to \textit{also} which the CLaC parser tagged as a DC signaling a \textsc{conjunction} relation. By projecting this annotation, we induce that \textit{aussi} should also be used in discourse usage and signals a \textsc{conjunction} relation. On the other hand, in Sentence~(\ref{ex:3}), \textit{aussi} is translated to \textit{both} which is not recognized as a DC, therefore, this French candidate DC is assumed to be used in a NDU. 

%%%%%%%%%%%%%%%%%%%%%%%%%%%%%%%%%%%%%%%%%%%%%%%%%%%%%%%%%%%%%%%%%%%%%%%%%%%%%%%%%%%%%%%%%%
\subsubsection{Unsupported French Candidate DCs}
\label{sec:unsupported}
%%%%%%%%%%%%%%%%%%%%%%%%%%%%%%%%%%%%%%%%%%%%%%%%%%%%%%%%%%%%%%%%%%%%%%%%%%%%%%%%%%%%%%%%%%
If the word-alignment model identified no alignments for a French candidate DC or aligned the candidate to punctuations, then we assume that the candidate has no translation and there is no annotation to be projected. We refer to such French candidate DCs as unsupported candidates and filter them before the annotation projection. Sentences~(\ref{ex:4}) and (\ref{ex:5}) in Table~\ref{tbl:annotation-projection} illustrate two cases of unsupported French candidate DCs. In Sentence~(\ref{ex:4}), the explicit French DC \textit{afin d'}\footnote{Free translation: \textit{in order to}} signals a \textsc{reason} relation, however it has been dropped in the English translation and replaced by the use of \textit{to + infinitive} (\textit{to assist}) to implicitly convey the \textsc{reason} relation. This example shows how the realization of DRs may be changed from explicit to implicit during the translation process. In Sentence (\ref{ex:5}), the French candidate DC \textit{pour}\footnote{Free translation: \textit{for}} does not signal a DR but again, it has no English translation. In both examples, since there is no English translation of the French candidate DCs, they will be filtered because there is no annotation that can be reliably projected onto them. 

Our approach is different from previous work as we identify unsupported French candidate DCs before the projection and filter them out. For example, \newcite{versley10} assumed that French candidate DCs are used in either a DU or a NDU. Anytime there is not enough evidence to label a French candidate DC as a DU (e.g. its translation is not part of an English DC), the candidate is assumed to be a NDU. This means that in Sentences (\ref{ex:3}), (\ref{ex:4}) and (\ref{ex:5}), all French candidate DCs would be tagged as NDU in \newcite{versley10}'s approach. On the other hand, our approach only labels the French candidate DC in Sentence (\ref{ex:3}) as NDU and filters out the French candidate DCs in Sentences (\ref{ex:4}) and (\ref{ex:5}) as they cannot be reliably annotated.

\refstepcounter{footnote}
\footnotetext{\label{fnote:examples}All examples are extracted from the Europarl corpus.}

%%%%%%%%%%%%%%%%%%%%%%%%%%%%%%%%%%%%%%%%%%%%%%%%%%%%%%%%%%%%%%%%%%%%%%%%%%%%%%%%%%%%%%%%%%
\subsection{Building the \corpus\ Corpora}
%%%%%%%%%%%%%%%%%%%%%%%%%%%%%%%%%%%%%%%%%%%%%%%%%%%%%%%%%%%%%%%%%%%%%%%%%%%%%%%%%%%%%%%%%%

Automatically aligning French candidate DCs to their English counterparts allowed us to automatically project discourse annotations from English onto French for each of the four word-alignment models. As a result, we created four different corpora from Europarl where French candidate DCs are labeled with either DU and the DR that they signal or NDU. We called these corpora: the \textit{\corpus\ corpora}\footnote{Available at \url{https://github.com/mjlaali/Europarl-ConcoDisco}.}. For comparative purposes, we also extracted a corpus without filtering unsupported candidates, which we refer to as \textit{Naive-Grow-diag}. Table~\ref{tab:stats} shows statistics of the corpora generated from Europarl. As the table shows, all corpora contain about 1 million French candidate DCs that are labelled as true French DC and for which a PDTB DR is assigned, and around 5 million candidates in non-discourse-usage. Compared to the FDTB, these corpora are approximately 100 times larger and French DCs are associated with PDTB relations. 

\begin{table}[htbp]
    \resizebox{\columnwidth}{!}{
    \rowcolors{2}{LightGray}{white}
    \begin{tabularx}{1.2\columnwidth}{|X|r|r|r|}
        \hline
        \rowcolor{Gray}
        \textbf{Corpus} & \multicolumn{1}{|l|}{\textbf{\# DU}} & \multicolumn{1}{|l|}{\textbf{\# NDU}} & \multicolumn{1}{|l|}{\textbf{Total}}
        \begin{filecontents*}{stats.csv}
file;Name;DU-raw;NDU-raw;Total-raw;DU;NDU;Total;order;ratio
intersection;Intersection;988,449;3,925,614;4,914,063;988;3,926;4,914;0;3.971488666
Grow-Diag;Grow-diag;1,074,306;5,191,079;6,265,385;1,074;5,191;6,265;1;4.832030166
en-fr;Direct;1,045,298;4,278,764;5,324,062;1,045;4,279;5,324;2;4.093343716
fr-en;Inverse;1,089,596;5,578,597;6,668,193;1,090;5,579;6,668;3;5.119876541
\end{filecontents*}
\csvreader[head to column names,/csv/separator=semicolon]{stats.csv}{}
        {\\\hline \corpus-\Name & \DU K & \NDU K & \Total K}
        \\ \hline Naive-Grow-diag & 1,090K & 5,191K & 6,265K
        \\ \hline 
    \end{tabularx}}
     \caption{Statistics of the \corpus\ and Naive-Grow-diag corpora.}
    \label{tab:stats}
\end{table}

As Table~\ref{tab:stats} shows, the \corpus\ corpora contain significantly different numbers of NDUs. For example, the \textit{Inverse} word-alignment model generates 1,653 thousands more NDU labels than the \textit{Intersection} word-alignment model (5,579K versus 3,926K). Section~\ref{sec:corpora-evaluation} discusses this difference and its relation to unsupported French candidate DCs.

%%%%%%%%%%%%%%%%%%%%%%%%%%%%%%%%%%%%%%%%%%%%%%%%%%%%%%%%%%%%%%%%%%%%%%%%%%%%%%%%%%%%%%%%%%
\section{Evaluation}
\label{sec:evaluation}
%%%%%%%%%%%%%%%%%%%%%%%%%%%%%%%%%%%%%%%%%%%%%%%%%%%%%%%%%%%%%%%%%%%%%%%%%%%%%%%%%%%%%%%%%%

%NEW
To evaluate our approach  to filtering unsupported annotations, we proceeded with two methods: 1) an intrinsic evaluation of both DU/NDU labels and the PDTB relations assigned to the French DCs in the \corpus\ corpora (see Section~\ref{sec:intrinsic-eval}) and 2) an extrinsic evaluation of DU/NDU labels using the task of disambiguation of French DC usage (see Section~\ref{sec:extrinsic-eval}).

%%%%%%%%%%%%%%%%%%%%%%%%%%%%%%%%%%%%%%%%%%%%%%%%%%%%%%%%%%%%%%%%%%%%%%%%%%%%%%%%%%%%%%%%%%
\subsection{Intrinsic Evaluation}
\label{sec:intrinsic-eval}
%%%%%%%%%%%%%%%%%%%%%%%%%%%%%%%%%%%%%%%%%%%%%%%%%%%%%%%%%%%%%%%%%%%%%%%%%%%%%%%%%%%%%%%%%%
To intrinsically evaluate the approach, we first built a gold-standard dataset using crowdsourcing (see Section~\ref{sec:gold-standard-dataset}), and then compared the \corpus\ corpora against the gold-standard dataset (see Section~\ref{sec:corpora-evaluation}).

%%%%%%%%%%%%%%%%%%%%%%%%%%%%%%%%%%%%%%%%%%%%%%%%%%%%%%%%%%%%%%%%%%%%%%%%%%%%%%%%%%%%%%%%%%
\subsubsection{Building a Gold-Standard Dataset}
\label{sec:gold-standard-dataset}
%%%%%%%%%%%%%%%%%%%%%%%%%%%%%%%%%%%%%%%%%%%%%%%%%%%%%%%%%%%%%%%%%%%%%%%%%%%%%%%%%%%%%%%%%%

To evaluate if French candidate DCs have the same discourse annotations as their translation, we designed a linguistic test, the \textit{Translatable} test, inspired by the \textit{Substitutability Test} of \newcite[p.~71]{knott96}. To investigate if two DCs signal the same relation, \newcite{knott96} compared a set of sentences where the only difference was the DCs used. If two sentences convey the same meaning then he assumed that the two DCs signal the same relation in that context. For example, the first two sentences in Example~\ref{ex:knott} (marked with a \checkmark) convey the same meaning, and therefore we can conclude that \textit{so} and \textit{thereby} signal the same relation in these two sentences. However, the third sentence (marked with a $\times$) does not convey the same meaning and therefore, it does not support that \textit{in short} can signal the same relation as the other two connectives\footnote{All sentences are taken from \cite{knott96}}. 
%We believe that the identification of the translation of French candidate DCs  we designed a test inspired by the \textit{Substitutability Test} of \newcite[p.~71]{knott96}. 

{\small
\begin{enumerate}[label=(\arabic*)]
    \setcounter{enumi}{\theexample}
    \refstepcounter{example}
    \item \label{ex:knott}
    \begin{itemize}
        \item[\checkmark] \textit{She left the country before the year was up; \textbf{so} she lost her right to permanent residence.}
        \item[\checkmark] \textit{She left the country before the year was up; she \textbf{thereby} lost her right to permanent residence.}
        \item[$\times$] \textit{She left the country before the year was up; \textbf{in short} she lost her right to permanent residence.}
    \end{itemize} 
\end{enumerate}
}

The \textit{Substitutability Test} has been also used by \newcite{roze12} as one of their linguistic tests to associate DRs to French DCs. 

%Our approach can be considered as a linguistic test to identify if the discourse annotation of an expression can be reliably projected to another expression in parallel texts. We refer to this test as the \textit{Translatable} test. 
Inspired by the \textit{Substitutability Test} test, we designed the \textit{Translatable} test. Since parallel sentences are a translation of each other, we can assume that they convey the same meaning and we therefore only need to verify if there is an English expression that is a good substitution for the French DC candidate. If this is the case, then we conclude that the French DC candidate should have the same discourse annotation (discourse usage and relation) as their English substitution. Otherwise, we conclude that the French DC candidate cannot be reliably annotated.

To build a gold-standard dataset, we first randomly selected  parallel sentences from a random Europarl file\footnote{ep-00-01-17.txt} containing French candidate DCs.
%To evaluate the corpora, 
%we performed an experiment in order to identify how often explicit French relations are change to implicit ones. More specifically, 
%we built a gold-standard dataset using crowdsourcing which consists of pairs of sentences that contain a French candidate DC and manually ran the \textit{Translatable} test on the dataset to build a gold-standard dataset. For the gold-standard dataset, the parallel sentences were randomly selected from a Europarl file\footnote{ep-00-01-17.txt} that contain French candidate DCs. 
For each French candidate DC, we selected at most 10 parallel sentences to keep the number of the sentence pairs tractable and to avoid any bias towards frequent French candidate DCs. This approach generated 696 pairs of parallel sentences similar to the examples in Table~\ref{tbl:annotation-projection}. Then, we used the CrowdFlower platform\footnote{\url{https://www.crowdflower.com/}} to run the \textit{Translatable} test on the dataset. To do so, we highlighted the French candidate DCs in each pair of parallel sentences (as shown in the column \textit{French} in Table~\ref{tbl:annotation-projection}) and asked annotators to identify (i.e. copy and paste) the English expression that is the best translation of the French candidate DC or to indicate if the French candidate DC has no translation. To ensure more accurate results, we limited the annotators to bilingual English-French speakers. Moreover, we manually aligned 80 test questions using three bilingual English-French speakers with a background in discourse analysis and filtered annotators whose accuracy was below 0.80 against these test questions. Out of 211 initial annotators, only 33 passed our test questions and proceeded with the actual annotation task. We used the webservice\footnote{\url{http://dfreelon.org/utils/recalfront/recal3/}} provided by \newcite{freelon10} to calculate the Krippendorff’s Alpha agreement \cite{krippendorff04} between the 33 annotators. The agreement between annotators was 0.787 which shows a strong agreement.

The CrowdFlower annotations allowed us to create a corpus of 696 pairs of sentences which we refer to it as the \textit{CrowdFlower gold-standard} dataset. Table~\ref{tab:sample-dataset} shows statistics of this  dataset. According to the crowdsourced annotators, 31.61\% of French candidate DCs can be substituted by an English DC which was marked by the CLaC parser and therefore are used in a DU (as in Sentence~(2) of Table~\ref{tbl:annotation-projection}); while 53.74\% can be substituted by an English expression which does not signal any DR according to the CLaC parser (as in Sentence~(3) of Table~\ref{tbl:annotation-projection}) and is therefore used in a NDU. Finally, 14.66\% of the French candidate DCs have no English translation (as in Sentences~(4) or (5) of Table~\ref{tbl:annotation-projection}), hence they cannot be reliably annotated. Recall that as opposed to previous work such as \cite{versley10}, our approach specifically addresses this significant proportion of explicit relations translated as implicit ones.

%raggedright/centered/raggedleft? 
\newcolumntype{L}[1]{>{\raggedright\let\newline\\\arraybackslash\hspace{0pt}}m{#1}}
\newcolumntype{C}[1]{>{\centering\let\newline\\\arraybackslash\hspace{0pt}}m{#1}}
\newcolumntype{R}[1]{>{\raggedleft\let\newline\\\arraybackslash\hspace{0pt}}m{#1}}
\newcolumntype{Z}{>{\raggedleft\let\newline\\\arraybackslash\hspace{0pt}}X}

\begin{table}[htbp]
    % \rowcolors{2}{LightGray}{white}
    \resizebox{1\columnwidth}{!}{
    \begin{tabularx}{1.3\columnwidth}{|r|r|r|Z|}
        \hline
        \rowcolor{Gray}
        \multicolumn{4}{|c|}{\textbf{French Candidate DCs}} \\ \hline
        \rowcolor{Gray}
        \multicolumn{1}{|c|}{\textbf{Total}} & \multicolumn{1}{|c|}{\textbf{DU}} & \multicolumn{1}{|c|}{\textbf{NDU}} & \multicolumn{1}{|c|}{\textbf{Dropped}}\\ \hline
         696 (100\%) & 220 (31.61\%) & 374 (53.74\%) & 102 (14.66\%) \\\hline 
         %(100\%) & (31.61\%) & (53.74\%)  & (14.66\%) \\ \hline
    \end{tabularx}}
    \caption{Statistics of the CrowdFlower gold-standard dataset.}
    \label{tab:sample-dataset}
\end{table}

%%%%%%%%%%%%%%%%%%%%%%%%%%%%%%%%%%%%%%%%%%%%%%%%%%%%%%%%%%%%%%%%%%%%%%%%%%%%%%%%%%%%%%%%%%
\subsubsection{Evaluation of the \corpus\ Corpora}
\label{sec:corpora-evaluation}
%%%%%%%%%%%%%%%%%%%%%%%%%%%%%%%%%%%%%%%%%%%%%%%%%%%%%%%%%%%%%%%%%%%%%%%%%%%%%%%%%%%%%%%%%% 

To evaluate the performance of the four word-alignment models in the identification of the English translation of French candidate DCs, we compared the corpora generated by the models against the CrowdFlower gold-standard dataset. Note that this evaluation shows the performance of the word-alignment models for the \textit{Translatable} Test, and therefore can be also considered as an intrinsic evaluation of the DRs assigned to the French candidate DCs\footnote{Because we do not have gold discourse annotations for Europarl, we can estimate the quality of the discourse annotations of the English side by evaluating the performance of the CLaC discourse parser on texts with a different domain such as the blind dataset of CoNLL shared task (see Section~\ref{sec:corpus-preparation}).}. Table~\ref{tab:word-alignments} shows precision (P) and recall (R) for both DU and NDU labels, as well as the overall annotations (OA) of the four \corpus\ corpora. As Table~\ref{tab:word-alignments} shows, the \corpus-Intersection  achieves the highest precision for both DU labels (0.934) and NDU labels (0.902), at the expense of recall. For example, while the \corpus-Intersection achieves a higher overall precision than the Naive-Grow-diag (0.914 versus 0.815), its recall is lower (0.845 versus 0.955).

\begin{table*}[htbp]
    \centering
    \rowcolors{2}{LightGray}{white}
    \resizebox{0.7\textwidth}{!}{
    \begin{tabularx}{0.8\textwidth}{|X|cc|cc|cc|}
         \hline
        \rowcolor{Gray}
         & \multicolumn{2}{c|}{\textbf{DU}} & \multicolumn{2}{c|}{\textbf{NDU}} & \multicolumn{2}{c|}{\textbf{OA}}\\ \cline{2-7}
         \rowcolor{Gray}
        \multirow{-2}{*}{\textbf{Corpus}} & \textbf{P} & \textbf{R} & \textbf{P} & \textbf{R} & \textbf{P} & \textbf{R} 
        \begin{filecontents*}{corpora-acc.csv}
Name,DUP,DUR,NDUP,NDUR,OAP,OAR
Intersection,0.934,0.895,0.902,0.816,0.914,0.845
Grow-diag,0.906,0.923,0.814,0.904,0.847,0.911
Direct,0.902,0.918,0.883,0.866,0.890,0.886
Inverse,0.891,0.927,0.801,0.928,0.832,0.928
\end{filecontents*}
\csvreader[head to column names]{corpora-acc.csv}{}
        {\\\hline \corpus-\Name & \DUP & \DUR & \NDUP & \NDUR & \OAP & \OAR}
        \\\hline Naive-Grow-diag & 0.906 & 0.923 & 0.771 & 0.973 & 0.815 & 0.955
        \\ \hline
    \end{tabularx}
    }
    \caption{Precision (P) and recall (R) of the four \corpus\ and the Naive-Grow-diag corpora against the CrowdFlower gold-standard dataset for DU/NDU labels and overall (OA).}
    \label{tab:word-alignments}
\end{table*}

Because the \textit{Intersection} model suffers from sparsity issues (many words are aligned to null), the \textit{Grow-diag} model is typically used for annotation projection \cite{tiedemann15,versley10}. However, Table~\ref{tab:word-alignments} shows that the \textit{Intersection} model is more suitable for discourse annotation projection due to its precision. Because the \corpus\ corpora are much larger than existing discourse corpora (with around 5 million annotations), a higher precision is preferable in our case.

A further error analysis shows that the main advantage of the \textit{Intersection} model is when French candidate DCs are dropped during the translation (i.e. explicit relations that are changed to implicit ones -- see the column \textit{Dropped} in Table~\ref{tab:sample-dataset}). For example in Sentence~\ref{ex:mais}, \textit{mais} has been dropped in the English translation. This causes both the \textit{Grow-diag} and the \textit{Inverse} models to incorrectly align \textit{mais} to \textit{and}. Hence, when we project the DR for either of these two models, \textit{mais} will be incorrectly marked as \textit{NDU} because \textit{and} is not an English DC. However, \textit{mais} signals a \textsc{contrast} relation. Therefore, a false-negative instance is generated for \textit{mais}. 

% \begin{enumerate}[label=(\arabic*)]
%     \setcounter{enumi}{\theexample}
%     \refstepcounter{example}
%     \label{ex:mais}
%     \item FR: \textit{\textbf{Mais} ma demande n'a pas été satisfaite.}\\
%     EN: \textit{That did not happen.}
% \end{enumerate}

Table~\ref{tab:dropped-dc} shows the performance of each alignment model for the identification of dropped French candidate DCs against the CrowdFlower gold-standard dataset. While the \textit{Intersection} model identifies the most dropped DCs (65\% out of the 102 dropped candidates), the \textit{Inverse} word alignment is the worst model as it identifies only 6\% of the dropped candidates and the naive \textit{Grow-diag} approach clearly identifies none. Note that the alignment models tend to label dropped French candidates DCs as NDU more often than as DU when they cannot identify candidates that were dropped during the translation; therefore, dropped French candidate DCs may artificially increase the number of NDU labels. This also explains why the number of NDU labels for the \textit{Intersection} word-alignment is the lowest among the word-alignment models (see Table~\ref{tab:stats}).

\begin{table}[htbp]
    \rowcolors{2}{LightGray}{white}
    \resizebox{1\columnwidth}{!}{
    \begin{tabularx}{1.2\columnwidth}{|X|r|cc|}
        \hline
        \rowcolor{Gray}
         &  & \multicolumn{2}{c|}{\textbf{Not identified}} \\ \rowcolor{Gray}
         &  & \multicolumn{2}{c|}{\textbf{and labeled as}} \\ \cline{3-4} \rowcolor{Gray}
         \multirow{-3}{*}{\textbf{Corpus}} & \multirow{-3}{*}{\textbf{Identified}} & \textbf{DU} & \textbf{NDU} 
         \begin{filecontents*}{dropped-dc.csv}
Name,None,NDU,DU
Intersection,65,29,8
Grow-diag,21,70,11
Direct,49,40,13
Inverse,6,79,17
\end{filecontents*}
\csvreader[head to column names]{dropped-dc.csv}{}
         {\\\hline \corpus-\Name & \None\% & \DU\% & \NDU\%}
         \\\hline Naive-Grow-diag & 0\% & 11\% & 89\%
         \\ \hline
    \end{tabularx}}
    \caption{Accuracy of the four \corpus\ and the Naive-Grow-diag corpora in the identification of dropped candidate DCs (unsupported candidates) against the CrowdFlower gold-standard dataset.}
    \label{tab:dropped-dc}
\end{table}

\subsection{Extrinsic Evaluation}
\label{sec:extrinsic-eval}
%%%%%%%%%%%%%%%%%%%%%%%%%%%%%%%%%%%%%%%%%%%%%%%%%%%%%%%%%%%%%%%%%%%%%%%%%%%%%%%%%%%%%%%%%%

%As shown in Section~\ref{sec:intrinsic-eval}, an intrinsic evaluation of the alignment models showed that the \textit{Intersect} model was the best model for discourse annotation projection. 
To extrinsically evaluate the effect of unsupported annotations on the quality of the \corpus\ corpora models, we used the corpora to train a binary classifier in order to detect the discourse usage of French DCs. Since the classifiers only differ by the training set used, by comparing the results of the classifiers, we indirectly assessed the quality of the corpora. 
%their differences in performance is directly attributed to the quality of their training datasets.

For our experiment, we used the French Discourse Treebank (FDTB) \cite{danlos15}. The FDTB marks French DCs in two syntactically annotated corpora: the Sequoia Treebank \cite{candito12} and the French Treebank (FTB) \cite{abeille00}. We assigned DU labels to the French DCs marked in the FDTB and NDU labels for all other non-discourse occurrences of the French DCs in the FDTB. Table~\ref{tab:fdtb-stats} shows statistics of the FDTB.

\begin{table}[htbp]
    \rowcolors{2}{LightGray}{white}
    \resizebox{1\columnwidth}{!}{
    \begin{tabularx}{1.2\columnwidth}{|X|r|r|r|}
        \hline
        \rowcolor{Gray}
        \textbf{Corpus} & \textbf{\# Word} & \textbf{\# DU} & \textbf{\# NDU} 
        \begin{filecontents*}{fdtb-stat.csv}
Sec;Word;DU;NDU
FTB;557,149;10,437;40,669
Sequoia;33,205;544;2,255
Total;579,243;10,735;42,924
\end{filecontents*}
\csvreader[head to column names,/csv/separator=semicolon]{fdtb-stat.csv}{}
        {\\\hline \Sec & \Word & \DU & \NDU}
        \\ \hline
    \end{tabularx}
    }
     \caption{Statistics of the FDTB. }
    \label{tab:fdtb-stats}
\end{table}

In our experiments, we used the same classifier used in the CLaC discourse parser \cite{laali16} for disambiguating the usage of English DCs and trained it on the \corpus\ corpora, the Naive-Grow-diag corpus and the FTB section of the FDTB. We reserved the Sequoia section of the FDTB for the evaluation of the trained classifiers. The text of the Sequoia section of the FDTB is extracted from Wikipedia and the ANNODIS corpus \cite{afantenos12}. This allowed us to compare the classifiers on datasets of different domains/genres than the training datasets, therefore, introducing no bias toward any of the training datasets. 
%We also defined a simple baseline classifier that labels every French DC with its most frequent class (i.e. \textit{DU} or \textit{NDU}) according to the FTB section of the FDTB.

Table~\ref{tab:results} shows the precision, recall and the F1-score of the classifiers.  While the precision of classifiers trained on the \corpus\ corpora is high (0.831\textasciitilde0.857) and actually higher than the one trained on the manually annotated FTB, their recall is much lower (0.309\textasciitilde0.406). We also observed that the classifiers trained on Naive-Grow-diag and on \corpus-Grow-diag have the same performance. This is because the Grow-diag models created many false-negative instances for a set of French DCs. Hence, the classifiers trained on this model labeled all occurrence of these French DCs as NDU. In addition, Naive-Grow-diag also added more false-negative instances to the same set of French DCs so the classifier labeled all those French DCs as NDU.

Among the classifiers trained on the \corpus\ corpora, the one based on the \textit{Intersection} model again achieves the best performance with an F1-score of 0.546. This confirms that the trade-off between precision and recall achieved by the \textit{Intersection} model makes it the most appropriate for discourse annotation projection.

The low recall of the classifiers trained on the \corpus\ corpora is an indication of a large number of false-negative instances. %: incorrect NDU labels which should have been labeled as DU.
%The false-negative instances in the datasets cause the classifiers to create more false-negative labels. 
As discussed in Section~\ref{sec:corpora-evaluation}, an important source of false-negative instances is due to French candidate DCs that are dropped in the translation. 
%Considering that actual French DCs are often dropped during the translation \cite{meyer13,zufferey12}, these results suggest that the dropped French candidate DCs are mostly true French DCs\footnote{Otherwise the dropped French candidate DCs would not generate false-negative instances.}. 
Table~\ref{tab:results} shows this by illustrating the same behaviour as in Table~\ref{tab:dropped-dc}. As these two tables show, the more accurate a word alignment model is at pruning dropped French candidate DCs, the higher recall the classifier will achieve using the dataset extracted from this word alignment model. In our case, the \textit{Intersection} model is the most accurate model in the identification of dropped candidate DCs with an accuracy of 65\% (see Table~\ref{tab:dropped-dc}), and the classifier trained on the \corpus-Intersection also achieves the highest recall (i.e. 0.406). This classifier achieves a 15\% relative improvement in F1-score compare to the one that was trained on Naive-Grow-diag. This shows the adverse effect of unsupported annotations on the classifiers.

To investigate further the low recall of the classifiers, we manually analyzed the results of three French DCs with a low recall and a high frequency in the CrowdFlower gold-standard dataset: \textit{enfin}, \textit{afin de} and \textit{ainsi}\footnote{Free translation: \textit{enfin} = \textit{finally}, \textit{afin de} = \textit{in order to}, \textit{ainsi} = \textit{so}.}. We observed that while 96\% of the French candidate DCs for these English DCs were properly aligned to their translation, 59\% of them were incorrectly labeled as NDU because their English translation were not properly annotated. This happened for three main reasons:
\begin{enumerate}
    \item The English translation is an English DC, but because it is either infrequent in the PDTB (e.g. \textit{finally}) or its NDU usage dominates its DU usage (e.g. \textit{for}), the English DC cannot be reliably annotated.
    \item The English translation is an English DC, but it is not listed in the PDTB (e.g. \textit{in order to}).
    \item The English translation is not an English DC, but it signals a DRs (e.g. \textit{this would ensure that} or \textit{in this way}). Such expressions are called \textit{AltLex} in the PDTB. We excluded AltLex from our analysis because to our knowledge, no English discourse parser can currently annotate them reliably.
\end{enumerate}

%Another reason for the low recall of the classifiers learned on the \corpus\ corpora can be that the annotations of the \corpus\ corpora are not representative for the annotations of the FDTB. The differences between the PDTB and the FDTB discourse framework or the differences between the genres of Europarl and Wikipedia/ANNODIS may cause this low recall\footnote{Note that while Europarl is composed of transcripts of speeches, Wikipedia/ANNODIS are written texts.}. For example, the translations of many French DCs marked in the FDTB are not considered in the PDTB and were categorized as AltLex, lexical or phrasal elements that are outside the initial set of the PDTB DCs. We excluded AltLex from our analysis because there is no English discourse parser that can reliably mark them and also since the list of AltLex is not considered as a closed list, to best of our knowledge, there is not a comprehensive inventory of AltLex. \todo{provide an example}  

\begin{table}[htbp]
    \resizebox{\columnwidth}{!}{
    \rowcolors{2}{LightGray}{white}
    \begin{tabularx}{1.2\columnwidth}{|X|cc|c|}
        \hline
        \rowcolor{Gray}
        \textbf{Training Corpus} & \textbf{P} & \textbf{R} & \textbf{F1} 
        \begin{filecontents*}{results.csv}
Dataset,P,R,F
FTB,0.777,0.756,0.766
\corpus-Intersection,0.831,0.406,0.546
\corpus-Grow-diag,0.837,0.331,0.474
\corpus-Direct,0.834,0.397,0.538
\corpus-Inverse,0.857,0.309,0.454
\end{filecontents*}
\csvreader[head to column names,/csv/separator=comma]{results.csv}{}
        {\\\hline \Dataset & \P & \R & \F }
        \\\hline Naive-Grow-diag & 0.837 & 0.331 & 0.474
        \\ \hline
    \end{tabularx}
    }
     \caption{Performance of the classifiers trained on different corpora against the Sequoia test set.}
    \label{tab:results}
        
\end{table}

%%%%%%%%%%%%%%%%%%%%%%%%%%%%%%%%%%%%%%%%%%%%%%%%%%%%%%%%%%%%%%%%%%%%%%%%%%%%%%%%%%%%%%%%%%
\section{Conclusion and Future Work}
%%%%%%%%%%%%%%%%%%%%%%%%%%%%%%%%%%%%%%%%%%%%%%%%%%%%%%%%%%%%%%%%%%%%%%%%%%%%%%%%%%%%%%%%%%
In this paper, we addressed the main assumption of annotation projection and showed that discourse annotations may not always be reliably projected in parallel sentences when DRs are changed from explicit to implicit ones during the translation. We proposed a novel approach based on the intersection between statistical word-alignment models to identify unsupported annotations. This approach was able to identify 65\% of the unsupported annotations, hence allowing the automatic induction of more precise corpora. As a by-product of our approach, we automatically induced the \corpus\ corpora: the first PDTB style discourse corpora for French. We showed that our approach to filtering unsupported annotations improves the F1-score of a classifier that labels the DU and the NDU of French DCs by 15\% compared to when the unsupported annotations are not filtered.   

There are several ways that this work can be extended. First, our method to induce a classifier to label French DCs with DU/NDU labels lends itself well to a bootstrapping approach. As we used English DCs to label the usage of French DCs, we could also use French DCs to label the usage of English DCs. Second, our approach can be used to automatically identify and annotate implicit DRs within English texts without parsing the English texts by identifying French DCs that are dropped during the translation (see Example~(\ref{ex:mais}) or Example~(\ref{ex:4})). In addition, since our approach only needs the availability of a parallel corpus with English, it can be easy used for other languages. Finally, the results of our work can be used to improve the development of French discourse resources such as LEXCONN and the FDTB.

\subsubsection* {Acknowledgement}
The authors would like to thank the anonymous referees for their insightful
comments on an earlier version of the paper. Many thanks also to Andre Cianflone, Alexis Grondin, Andrés Lou and Félix-Herve Bachand for their help on the CrowdFlower task. This work was financially supported by an NSERC grant.

\bibliographystyle{acl_natbib}

\end{document}